# From Uncertainty to Failure Attribution:
# Self-Diagnosing Models for Failure Attribution under Distribution Shift

**Yiyao Yang**
**Columbia University**
**ORCID: 0009-0001-8693-4888**
**yy3555@tc.columbia.edu**

**Abstract**

Distribution shift poses a significant challenge to the robustness of machine learning models, but the current solutions only aim to detect out-of-distribution (OOD) samples and predict uncertainty levels. We introduce a problem setting for failure attribution under distribution shift, which enables the models not only to detect OOD samples, but also to find out the reason for their failure. The solution we propose is called self-diagnosing models, which are capable of jointly learning predictive output, predictive uncertainty, and a failure attribution signal. In particular, we use the failure attribution vector, produced by a neural network, which provides a structured representation of predictive unreliability by distinguishing four different types of failures: covariance shift, semantic shift, noise corruption, and adversarial perturbation. In other words, we move from scalar uncertainty towards failure identification. For training the model, we introduce a consistency regularizer that encourages consistency between uncertainty and failure attribution predictions. Moreover, to be able to evaluate the model on its ability to find the reasons for failure, we construct several distribution shift benchmarks with predefined mechanisms for generating distribution shifts.

## I. Introduction

Despite remarkable progress made by modern machine learning models on various tasks, robustness under distribution shift remains an open and important challenge. The presence of covariate shifts, semantic shifts, corruptions, or attacks may lead to severe reductions in predictive accuracy (Quiñonero-Candela, 2009; Hendrycks & Dietterich, 2019). Such problems become particularly problematic in application scenarios with high stakes, as an erroneous prediction combined with high model confidence can result in a suboptimal decision. Much previous research has approached this problem via uncertainty estimation and OOD detection. In the first case, various Bayesian approximation techniques, including Monte-Carlo dropout (MC dropout), are commonly used to provide tractable uncertainty estimates for predictions (Gal & Ghahramani, 2015). Recently, deep ensembles were proposed as an effective uncertainty estimator that scales well to modern deep learning models (Lakshminarayanan et al., 2016; Ovadia et al., 2019; Wilson & Izmailov, 2020). In the second case, OOD detection methods attempt to discriminate between in-distribution and out-of-distribution examples. Initially, confidence-based approaches like simple classification confidence thresholding were developed (Hendrycks & Gimpel, 2018), but then more complex techniques such as temperature scaling or input transformations (Liang et al., 2020), Mahalanobis distance in feature space (Lee et al., 2018), outlier exposure (Hendrycks et al., 2019), and energy-based scores (Liu et al., 2020) have been introduced.

The vast majority of proposed approaches tackle a narrowly defined task, determining when a model's predictions should be trusted. They produce scalar measures, such as uncertainty scores, predictive entropy, or OOD scores, which indicate abnormality or lack of trustworthiness without any explanation as to the type of problem. Even in the case where those approaches successfully identify problematic cases, they tend not to distinguish between the types of problems leading to unreliable performance, such as covariate shift, semantic gap, data corruption, adversarial manipulation, etc. (Hendrycks & Gimpel, 2018; Lee et al., 2018; Liu et al., 2020; Yang et al., 2021). Therefore, these methods cannot aid in diagnosing what exactly happens inside the network and how its performance could be improved. Nevertheless, in practice, knowing what kind of problem a model encounters is equally important as being able to determine if a particular example could be potentially problematic. In fact, different failure modes might require different actions on behalf of developers (adaptation/reweighting in the case of covariate shift, denoising or robustifying for corrupted images, etc.) (Quinonero-Candela et al., 2009). Therefore, instead of defining reliability as uncertainty estimation, we believe that it should be understood as failure attribution. Namely, a truly reliable model should attribute predictive unreliability to interpretable failure modes reflecting different types of problems. This claim is further supported by recent findings that showed how the quality of uncertainty estimates can deteriorate due to shifts in distribution despite good in-distribution performance (Ovadia et al., 2019).

We explore the problem of failure attribution in the case of distribution shifts. While conventional OOD research focuses on the problem of recognizing the failure, we are concerned with not only detecting when a model fails but also understanding why. To do this, we introduce the concept of self-diagnosing models as a general approach capable of learning predictive output, confidence score (uncertainty), and structured failure attribution simultaneously. Specifically, a self-diagnosing model generates a failure attribution vector that identifies various sources of failure as separate dimensions: covariate shift, semantic shift, noise corruption, and adversarial perturbation. In order to facilitate this shift from simple uncertainty analysis to more mechanism-aware failure diagnosis, we develop an MTL framework with a consistency regularizer that encourages alignment of uncertainty estimation with failure attribution. In addition, we introduce a benchmark for failure attributions with controlled distribution shift, based on a synthetic generative process for which we can directly evaluate whether failure attributions generated by a model accurately capture the underlying causality. Such evaluations are essential due to the fact that common OOD benchmarks focus primarily on detection and do not have intrinsic mechanisms for evaluating causal attribution (Hendrycks & Dietterich, 2019; Liu et al., 2020; Yang et al., 2021).

The contributions include defining failure attribution for distribution shifts as a novel learning problem, going beyond traditional uncertainty modeling and OOD detection, developing self-diagnosing models that learn prediction, uncertainty, and failure attribution simultaneously, and devising controlled evaluation procedures, demonstrating that model failures can be diagnosed, and explaining how such an approach is crucial for building robust machine learning systems.

## II. Research Questions

**Research Question 1:** Can models learn not only when predictions are unreliable, but also why they fail under distribution shift?
**Research Question 2:** Does failure attribution improve reliability and interpretability beyond standard uncertainty-based methods?

## III. Result

This study uses the CIFAR-10 dataset (Krizhevsky, A., et al), which contains 60,000 color images of size 32×32 across 10 classes, with 6,000 images per class. The dataset is divided into 50,000 training images and 10,000 test images. The programming language used in this study is Python.

Given an input image $x$, the proposed self-diagnosing model first learns a shared latent representation $h = \phi(x)$. Based on this shared representation, the model jointly performs three tasks, including class prediction, reliability estimation, and failure attribution. The classification head outputs $p^{(c)} = \text{softmax}(W_c h + b_c)$, and the reliability head outputs $p^{(r)} = \text{softmax}(W_r h + b_r)$. To enhance failure attribution, we further construct representation-space diagnostic features $d = [p^{(c)}_{(1)}, p^{(c)}_{(1)} - p^{(c)}_{(2)}, ||h||_2]$, where $p^{(c)}_{(1)}$ and $p^{(c)}_{(2)}$ denote the top-1 and top-2 class probabilities. The failure attribution head then predicts the failure type based on the concatenated feature vector $[h; d]$, yielding $p^f = \text{softmax}(g([h; d]))$. The entire model is trained end-to-end with a weighted multi-task objective $\mathscr{L} = \mathscr{L}_{\text{classification}} + 0.7\mathscr{L}_{\text{failure attribution}} + 0.5\mathscr{L}_{\text{reliability}}$.
$\mathscr{L}_{\text{classification}} = \text{CE}(z^{(c)}, y)$, where $z^{(c)}$ denotes the classification logits, and $y$ is the ground-truth class label in CIFAR-10. $\mathscr{L}_{\text{failure attribution}} = \text{CE}(z^{(f)}, f)$, where $z^{(f)}$ denotes the failure attribution logits, and $f$ is the true failure label. $\mathscr{L}_{\text{reliability}} = \text{CE}(z^{(r)}, r)$, where $z^{(r)}$ denotes the reliability logits, and $r$ is the reliability label: $r = 0$ indicates reliable and $r = 1$ indicates unreliable.

**Research Question 1**

In our first formulation of semantic shift, semantic shift was achieved via perturbations within the label space, where there were no guarantees that these led to perceptible changes within the input distribution itself. While useful for showcasing theoretical limits, such an approach makes it hard to detect the presence of a semantic shift using only the input space. To solve this, we modify our semantic shift formulation to include a degree of structural perturbations alongside semantically close replacements of class labels. This approach allows us to preserve many of the lower-level characteristics but alter their corresponding alignments within the label space, resulting in mismatches that could be evident in the learned feature space. We additionally include confidence, margin, and feature norm measures in the representation space to allow for failure attribution in the semantic shift formulation.

In order to investigate Research Question 1, we conducted experiments to verify if the model is able to learn not only when but also why the model fails because of distribution shift. The test results showed that the model obtained a classification accuracy of 0.5983, a failure attribution accuracy of 0.9433, a reliability detection accuracy of 0.9440, and a reliability AUROC of 0.9864. In other words, the model is able to jointly learn classification, reliability, and failure attribution. Mathematically speaking, the model learns three outputs for an input $x$: $\hat{y} = f(x)$, $\hat{r} = g(x)$, and $\hat{a} = h(x)$, where $\hat{y}$ denotes the predicted label, $\hat{r} \in [0,1]$ represents the predicted unreliability score, and $\hat{a} \in R^K$ is the failure attribution vector over $K$ shift types. The predicted failure type is given by $\hat{k} = argmax_j \hat{a}_j$, and the overall failure attribution accuracy is defined as Attribution Accuracy = $P(\hat{k} = k)$. For identifying when predictions are unreliable, the model showed reasonably good overall performance. We define prediction unreliability as $r = \mathbb{I}(\hat{y} \neq y)$, and train the model to approximate this indicator via a learned function $\hat{r} = g(x)$. From the perspective of the joint error analysis (overall discrimination), AUROC reached 0.9864, indicating an excellent ability to distinguish reliable from unreliable samples. However, this ability was less effective at the individual case level. For example, the number of predictions that were inaccurately made but recognized as erroneous by the system stood at 0.6609.

$$\text{Error Awareness Rate (EAR)} = \frac{\sum_i \mathbb{I}(\hat{y}_i \neq y_i \ \wedge \ \hat{r}_i = 1)}{\sum_i \mathbb{I}(\hat{y}_i \neq y_i)} = 0.6609$$

In addition, only 0.4165 of the correct predictions were labeled as reliable. This suggests that the model learned broad patterns of unreliability, but its self-awareness was still incomplete.

$$\text{Correct Confidence Rate(CCR)} = \frac{\sum_i \mathbb{I}(\hat{y}_i = y_i \ \wedge \ \hat{r}_i = 0)}{\sum_i \mathbb{I}(\hat{y}_i = y_i)} = 0.4165$$

To account for the reasons behind the failures, there were significant differences between the types of shifts. Failure attribution performance was very high under the clean, covariate, noisy-shifts, and semantic shift scenarios. Its values reached 0.9115, 0.9169, 0.9990, and 0.9813, respectively. The performance of the reliability task was also impressive, suggesting that the model performed well in identifying failures caused by lower-order perturbations such as changes in brightness and noise levels. Semantic shift is no longer difficult to distinguish from clean samples. The new architecture allows for an accuracy rate of 0.4682 in classification, failure attribution accuracy of 0.9813, and reliability accuracy of 0.9851 under semantic shift. In essence, samples undergoing semantic shifts are recognizable and easily separable from the original samples instead of being classified.

The failure attribution confusion matrix for all four types of shifts (Figure 1) shows that the proposed model successfully makes attributions for most of the clean, covariate, and noise failure cases, identifying 3,648 out of 4,002 clean, 1,942 out of 2,118 covariate, and 2,005 out of 2,007 noise failure samples. Semantic shifts can be attributed accurately. Indeed, 1,838 out of 1,873 semantic samples are attributed correctly to the semantic category, whereas 33 are incorrectly attributed to the clean category. Thus, we can conclude that the model is able to attribute not only low-level failures associated with observable perturbation effects but also high-level failures resulting from semantic shifts.

From the normalized failure attribution confusion matrix for the shifted failure attribution model (Figure 2), it can be seen that the proposed model successfully detects the shift at the low end of the distribution with accuracies of 91%, 92%, and 100% in clean, covariate, and noisy samples, respectively. Semantic failures that were not detectable using the previous model can now be detected using the proposed model with an accuracy of 98%.

**Figure 1**
*Failure Attribution Confusion Matrix*

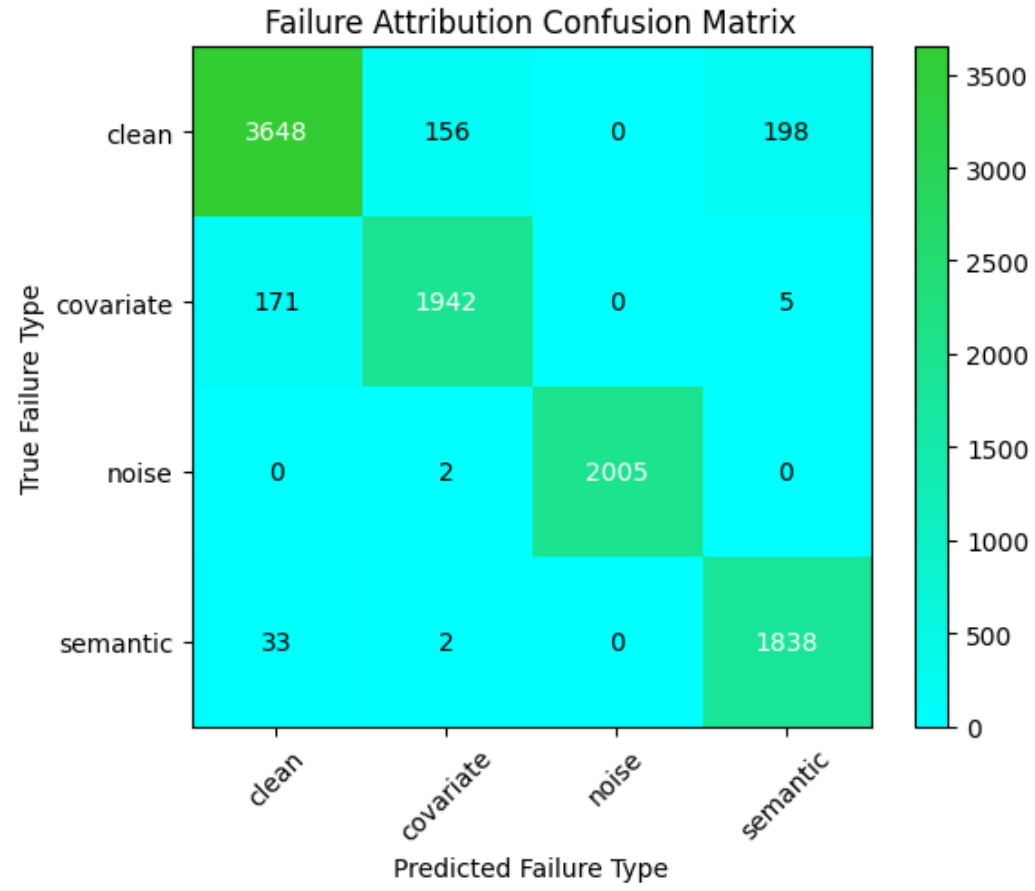

**Figure 2**
*Normalized Failure Attribution Matrix*

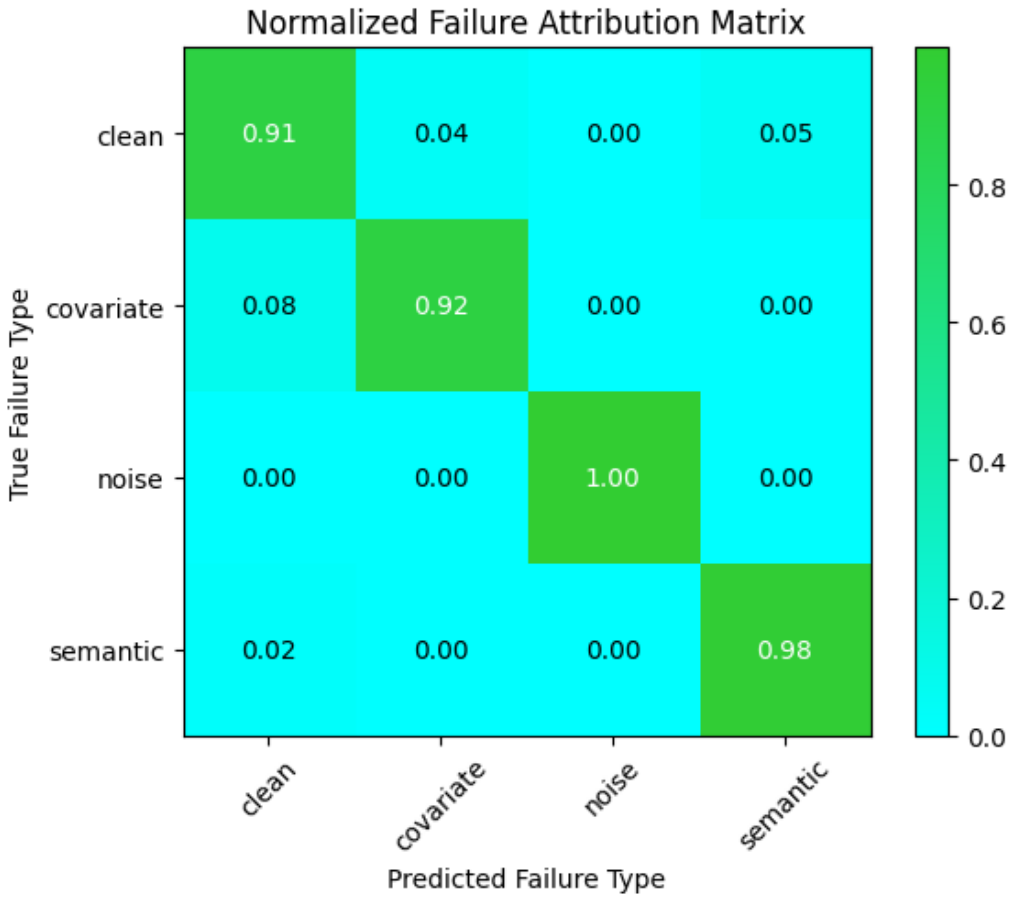

Based on the reliability analysis (Table 1), the model obtains impressive values of reliability accuracy of 0.94, as well as an AUROC of 0.99, suggesting its capability of distinguishing between reliable and unreliable cases across the entire dataset. Nevertheless, a more detailed investigation of the confusion matrix shows a critical deficiency in the model's functioning. There are 206 samples of unreliable cases that have been incorrectly predicted by the model as reliable cases. This means that the model cannot determine whether certain predictions are erroneous or not and thus cannot provide an appropriate reliability indicator. Since the confusion matrix is $\begin{bmatrix} 3648 & 354 \\ 206 & 5792 \end{bmatrix}$, the EAR is calculated as $\mathrm{EAR} = \frac{5792}{5792 + 206} = 0.9657$, suggesting that only 96.57% of the model's errors are identified as unreliable, while 3.43% of them go unnoticed. Despite learning some valuable global features of uncertainty, the model faces difficulties when recognizing some errors in predictions, due to the fact that they are connected to high-confidence or ambiguous representations.

**Table 1**
*Reliability Report*

| | Precision | Recall | F1-Score | Support |
|---|---|---|---|---|
| Reliable | 0.95 | 0.91 | 0.93 | 4002 |
| Unreliable | 0.94 | 0.97 | 0.95 | 5998 |
| Accuracy | - | - | 0.94 | 10000 |
| Macro Avg | 0.94 | 0.94 | 0.94 | 10000 |
| Weighted Avg | 0.95 | 0.94 | 0.94 | 10000 |

From the joint error analysis, the model possesses an average yet somewhat incomplete capability for self-reflection. Given the total number of prediction errors made by the model at 4,017, the model was able to identify 2,655 errors accurately as being inaccurate, hence yielding an error awareness rate of 0.66. This would imply that the model has an above-average capability of recognizing its errors, albeit with some errors still not being recognized. On the other hand, regarding the analysis of the failure reason, the model has performed excellently. From the total number of errors made by the model, 3,754 of them have been attributed to the right failure reason, thereby producing a reason identification rate of 0.93. Error-Case Attribution Accuracy $= \mathbb{P}(\hat{k} = k \mid \hat{y} \neq y)$.

Classification by the network does not demonstrate good classification accuracy in semantic shift cases; thus, we can assume that semantic problems are still more difficult to overcome than perturbations at lower levels. In the Latent Diagnostic Map (Figure 3), the representation of noise perturbation forms a well-separated compact cluster located in the top part of the representation space. At the same time, covariate shift cases appear in the center part of the space and partially overlap with clean data points. Finally, semantic shift cases appear mostly in the bottom part of the space and are grouped into a compact cluster, yet with a higher degree of overlap with other failure types.

Based on the Confidence-Unreliability Phase Map (Figure 4), there is a clear distinction between reliable and unreliable samples. Reliable samples appear to be clustered in the low-unreliability phase space area, where many samples demonstrate low levels of unreliability probability at various levels of classification confidence. On the contrary, noise-affected samples are located primarily in the high-unreliability area, with some exhibiting extremely high probabilities of unreliability, almost reaching unity; hence, it is easier for the model to detect the presence of noise. Covariate-shifted and semantic-shifted samples are mostly found in the high-unreliability phase space area, which means that the model is able to detect both low-level and structured distribution shifts. Nevertheless, samples subjected to a semantic shift are more widely dispersed along the confidence axis, with many having high confidence and low reliability.

**Figure 3**
*Latent Diagnostic Map*

**Figure 4**
*Confidence-Unreliability Phase Map*

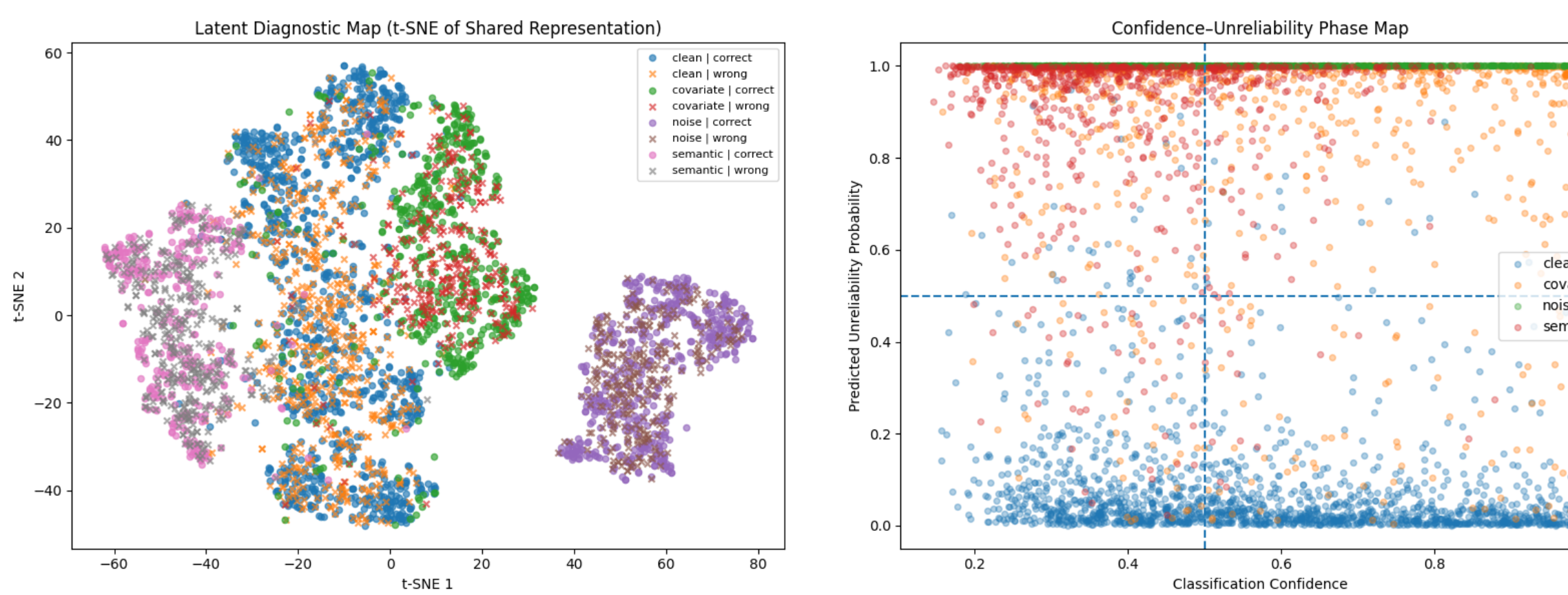

The intuition behind the different types of distribution shifts considered in this study (Figure 5) presents representative examples of clean, covariate-shifted, noise-corrupted, and semantic-shifted samples. Whereas covariate and noise shift predominantly cause perturbations at a relatively low level, semantic shift under the new framework results in mismatches that can be detected in the learned feature space, thereby making such examples easier to detect compared to the previous case. Based on the Illustration of a Failed Semantic Shift (Figure 6), even though incorrect class predictions continue to be made on examples affected by a semantic shift, most of them are correctly classified as belonging to the semantic failure cluster (fail: 3), as well as being marked as unreliable (rel: 1).

**Figure 5**
*Sample Gallery of Four Types of Shifts*

**Figure 6**
*Illustration of a Failed Semantic Shift*

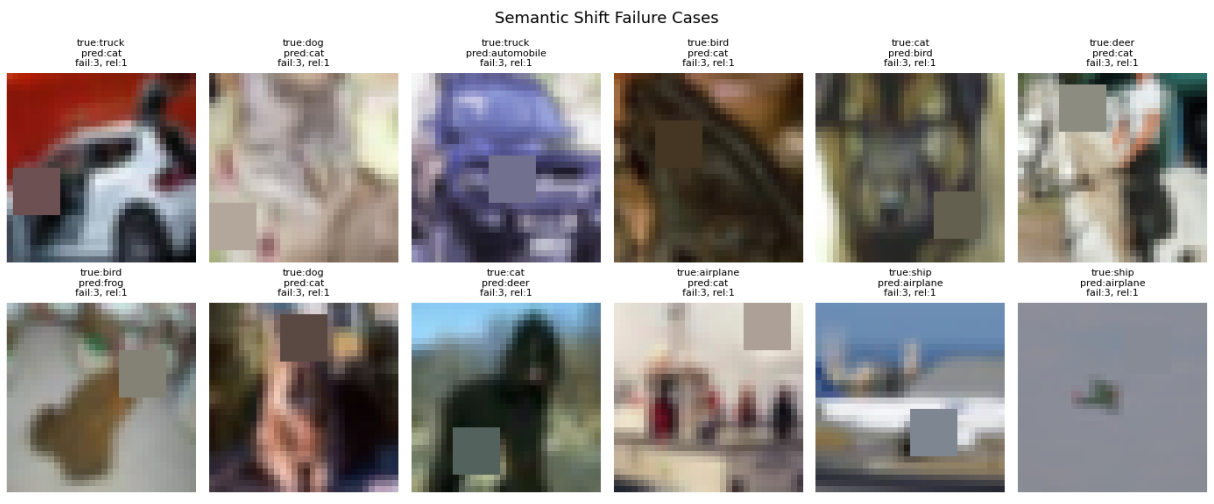

The identifiability problem of the original semantic shift can be summarized (Appendix VI.1, Proposition 1). Denote $X$ as the observable input, and $S \in \{0, 1\}$ represents a binary semantic shift. Assume that the semantic shift does not change the probability of the input $P(X \mid S = 1) = P(X \mid S = 0)$. Then, for all measurable decision rules $h$, there is no classifier based only on input $X$, that is better than random guessing with equal prior probabilities.

The high-confidence error case is illustrated (Figure 7), where the model outputs wrong classifications with a very high degree of confidence (confidence ~ 0.95 – 0.99). These examples show one major drawback of the model. While it performs reliably in general, there is a possibility that it can give a wrong prediction even at high confidence. Therefore, some errors, especially those made by high-confidence representations, cannot be easily detected by the model. However, based on the High-Confidence Wrong Prediction, the output prediction is wrong and extremely confident (confidence ≈ 0.95 – 0.99). Although the model performs reliably when detecting unreliability, it also has a drawback whereby it makes highly confident, wrong predictions. In other words, mistakes based on high-confidence representations are harder to detect. The high-confidence error is formally described as High-Confidence Error $= P(\hat{y} \neq y \wedge \max_j p_j(x) > \tau)$, where $p_j(x)$ denotes the predicted probability for class $j$, and $\tau$ is a confidence threshold.

**Figure 7**
*High-Confidence Wrong Predictions*

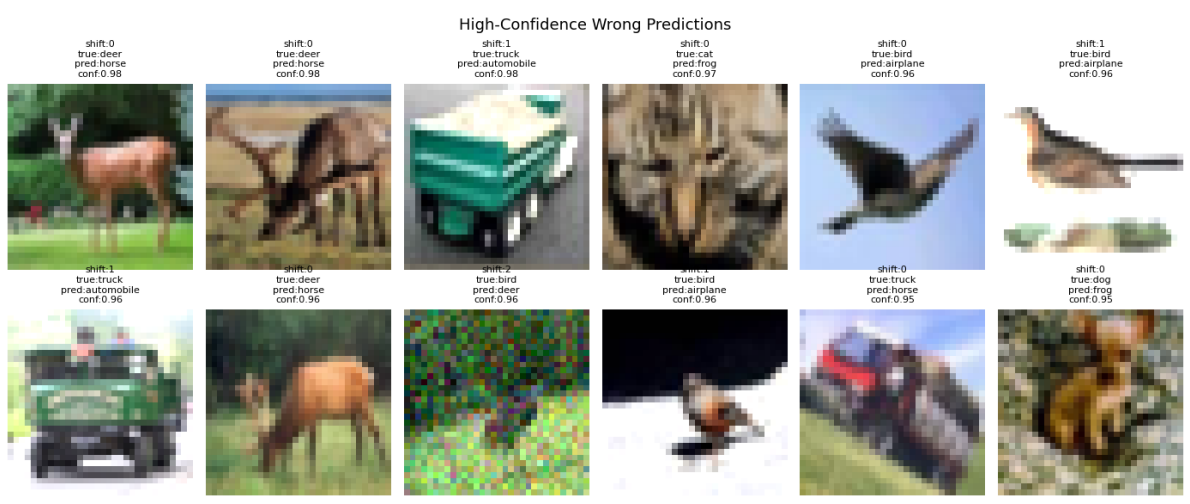

As observed from the experimental results, the proposed model is highly effective in terms of diagnostics under a distribution shift scenario, being able not only to detect the cases where failures occur but also to determine the type of failure that occurs. Under low-level distribution shifts like covariate and noise shifts, the model is especially efficient in diagnosing the type of failures occurring within each dataset. The results obtained under the semantic distribution shift show some differences when compared with those obtained under other types of distribution shifts. Though there is relatively lower classification accuracy under the semantic shift, the model is able to identify any failures occurring and categorize them under the semantic failure type. Nevertheless, the occurrence of highly confident mistakes proves that there is some insufficiency in the self-diagnostic capability of the model. In particular, certain representation-related mistakes do not get detected as unreliable.

**Research Question 2**

Before presenting the empirical results, we provide a theoretical justification for why incorporating failure attribution can improve reliability estimation beyond standard uncertainty-based methods. (Appendix VI.2, Proposition 2).

In order to determine how useful the introduction of failure attributability is for the diagnosis process, a comparison of the proposed model was made with respect to some well-established benchmark models, such as the Uncertainty-only baseline model, MC Dropout, Deep Ensemble, and the Energy Score model.

From the above comparison table (Table 2), it can be observed that the Uncertainty-only Baseline model has an outstanding result in terms of reliability, with an accuracy of 0.9290 and an AUROC of 0.9796. This implies that traditional uncertainty features have abundant information to distinguish between reliable and unreliable data points. Nevertheless, by integrating the uncertainty signals with the failure attribution approach, we can further enhance the reliability evaluation, as our model has obtained an accuracy of 0.9440 and an AUROC of 0.9864 in the reliability assessment task. In the error-awareness task, although the Uncertainty-only Baseline model obtains a slightly higher EAR score of 0.6622 compared to our result (0.6609), it fails to provide the ability to detect the root cause of the failures. On the other hand, in our model, we obtain the best performance in terms of accuracy and AUROC of reliability evaluation while simultaneously providing the failure attribution. Besides, our approach provides the highest accuracy of 0.9433 in terms of failure attribution and 0.9345 in terms of reason accuracy of errors.

We also demonstrated mathematical definitions of all evaluation metrics (Appendix VI.3, Mathematics Formula of Evaluation Metrics).

**Table 2**
*Comparison Table*

| Model | Reliability Accuracy | Reliability AUROC | Error Awareness Rate (EAR) | Correct Confidence Rate (CCR) | Failure Attribution Accuracy | Failure Reason Identification Accuracy on Error Cases |
|---|---|---|---|---|---|---|
| Uncertainty-Only | 0.9290 | 0.9796 | 0.6622 | 0.4060 | - | - |
| MC Dropout | 0.4928 | 0.5786 | 0.4311 | 0.8526 | - | - |
| Deep Ensemble | 0.4884 | 0.5713 | 0.4453 | 0.8543 | - | - |
| Energy-Based Score | 0.4166 | 0.4414 | 0.0840 | 0.6564 | - | - |
| Self-Diagnosing + Failure Attribution | 0.9440 | 0.9864 | 0.6609 | 0.4165 | 0.9433 | 0.9345 |

Based on the findings (Table of Per-Shift Results, Appendix VI.4), when taking into account all types of distribution shift, it may be said that the hardest scenario for our problem statement is that of the semantic setting. The reason for this is the low accuracy of classification, which fluctuates from 0.4677 to 0.5302. However, when other scenarios apply (clean, covariate, and noise settings), the accuracy levels of classification are considerably high and oscillate from 0.5899 to 0.7054. Moreover, as far as the scenario of the semantic setting is concerned, it seems that the approaches based on uncertainty fail. For instance, the reliability accuracy of MC Dropout is 0.4640, while in the Deep Ensemble scenario, it equals 0.4383.

In particular, the Key Results of Semantic Shift (Table 3) show that the uncertainty-only model achieves a high reliability accuracy of 0.9722, as well as a high error awareness rate of 0.9511. However, the model shows a very poor correct confidence rate of 0.0091. As a result, it can be stated that this model is very reluctant to treat predictions as reliable, marking most cases as unreliable under semantic shift. In comparison, the reliability accuracy of the MC Dropout and Deep Ensemble models is considerably lower, reaching 0.4640 and 0.4383, respectively. At the same time, the correct confidence rates of the two approaches equal 0.6542 and 0.6918, respectively, indicating that these techniques are more capable of preserving confidence in correct predictions. However, their error awareness rates, 0.5970 and 0.5527, remain noticeably lower, which implies that they lack the ability to detect a substantial portion of failed cases under semantic shift. Finally, regarding the energy-based score approach, it provides the lowest rates among all other approaches discussed above. In particular, the reliability accuracy of the technique is only 0.0662, while the error awareness rate equals 0.0341. On the other hand, the model is able to detect correct cases with a relatively high probability, reflected by a correct confidence rate of 0.9053. As for our algorithm, it reaches a reliability accuracy of 0.9851 and an error awareness rate of 0.9809. Meanwhile, the correct confidence rate of our algorithm remains very low at 0.0103.

**Table 3**
*Key Results of Semantic Shift*

| Model | Reliability Accuracy | Error Awareness Rate | Correct Confidence Rate |
|---|---|---|---|
| Uncertainty-Only | 0.9722 | 0.9511 | 0.0091 |
| MC Dropout | 0.4640 | 0.5970 | 0.6542 |
| Deep Ensemble | 0.4383 | 0.5527 | 0.6918 |
| Energy-Based Score | 0.0662 | 0.0341 | 0.9053 |
| Self-Diagnosing + Failure Attribution | 0.9851 | 0.9809 | 0.0103 |

From the Diagnostic Evaluation of Self-Diagnosing Models under Distribution Shift (Figure 7), the Radial Diagnostic Wheel indicates that the proposed model performs satisfactorily. It shows that the proposed model possesses a high level of reliability accuracy (0.9440), reliability AUROC (0.9864), and failure attribution accuracy (0.9433). This implies that the model is capable of predicting the reliability of the prediction and attributing the cause of failure. As shown in the Semantic Shift: Method Comparison, the scenario of semantic shift proves to be the most difficult environment. The models of MC Dropout, Deep Ensemble, and the energy-based method are unable to handle the situation and obtain reliability accuracy of 0.4640, 0.4383, and 0.0662, respectively. The proposed model, however, is able to produce 0.9851 reliability accuracy and 0.9809 EAR in this condition. It appears that failure attribution has been effective in maintaining reliability detection capability in difficult shifts. The Failure Flow Map provides an insight into the interpretability advantage offered by the proposed model. It shows that clean, covariate, and noise shifts have been correctly identified, but the semantic shift is usually categorized as a clean shift. Such a clear distinction between different shifts allows people to understand the ability and limitations of the model. This type of interpretation is impossible using conventional uncertainty measurement techniques.

**Figure 8**

*Self-Diagnosing Models under Distribution Shift*

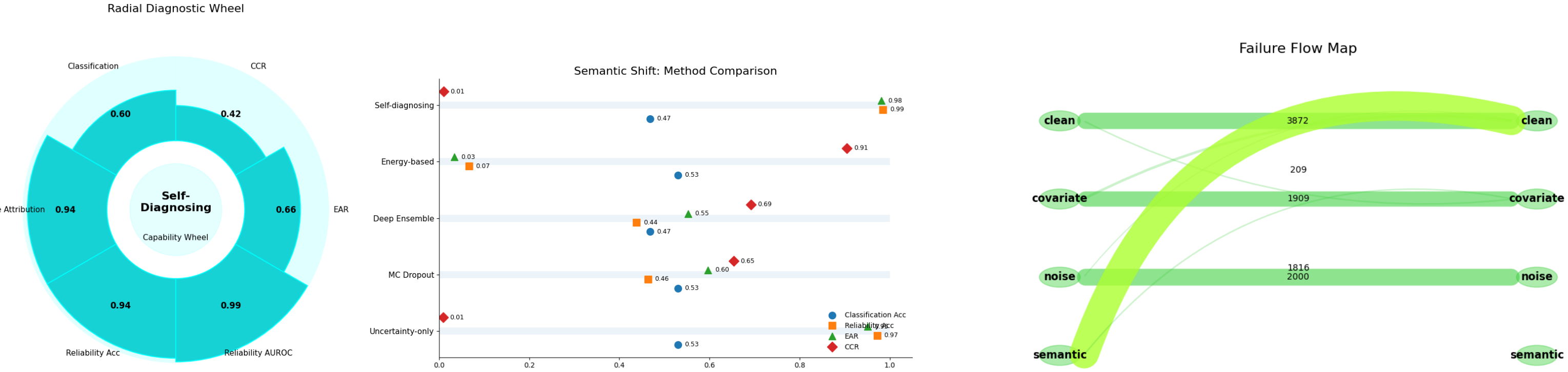

## IV. Conclusion

This paper investigated the issue of failure attribution in situations of distributional shift and developed a self-diagnosing system where prediction, reliability estimates, and explanatory failure reasons are learned simultaneously. In contrast to uncertainty-based approaches that are capable of determining only whether the prediction is reliable, our solution provides more information about the nature of failures, which include, among others, covariate shift, semantic shift, noisy input, and perturbative attacks. The experiments conducted on CIFAR-10 data showed the high efficiency of the model in terms of overall results and its high failure attribution accuracy, which amounted to 0.9429. The most effective in terms of detection was the performance on the problems of clean data, covariate shift, and noise. Our model has also been successful in determining the situation of semantic shift, performing better than the previous version of itself. When comparing with uncertainty-based models such as MC Dropout, Deep Ensemble, and Energy-based approaches, the results showed that the reliability estimation performed better than the previous approaches while explaining the failures in terms of failure attribution.

## V. Limitation

The existing approach considers four failure types, namely covariate shift, semantic shift, noise corruption, and adversarial perturbation. In practice, there might be cases where the failures could be a result of some other factors, which are not limited to those four categories. Therefore, for future research, it is recommended to consider a broader set of failures that could possibly occur, as well as different approaches to attributing them to models. Moreover, the reliability module is not fully capable of identifying all the mistakes in predictions individually. While the overall accuracy, reliability, and AUROC are impressive for the proposed system, some incorrect predictions have still escaped detection. This indicates that while the model can identify certain global unreliabilities, there are some cases where it lacks proper self-awareness.

## VI. Appendix

### (VI.1) Proposition 1: Identifiability under Semantic Shift

Suppose $\in \{0, 1\}$ indicates whether a semantic shift occurs. If $P(X \mid S = 1) = P(X \mid S = 0)$, then no classifier based solely on $X$ can distinguish between $S = 1$ and $S = 0$ better than chance under equal class priors. This result formalizes an identifiability limitation of semantic shift under observational equivalence.

**Proof:** Assume equal class priors, that is, $P(S = 1) = P(S = 0) = \frac{1}{2}$. By Bayes' rule, for any observable input $x$,

$$P(S=1 \mid X=x) = \frac{P(S=1 \cap X=x)}{P(X=x)} = \frac{P(X=x \cap S=1)}{P(X=x)} = \frac{P(X=x \mid S=1)P(S=1)}{P(X=x \mid S=1)P(S=1) + P(X=x \mid S=0)P(S=0)} = \frac{P(X=x \mid S=1)P(S=1)}{P(X=x \mid S=1) + P(X=x \mid S=0)P(S=0)}$$

. Since $P(S = 1) = P(S = 0) = \frac{1}{2}$ as the assumption, $P(X \mid S = 1) = P(X \mid S = 0)$.

Let $P(X \mid S = 1) = P(X \mid S = 0) = L$.

$$P(S = 1 \mid X = x) = \frac{P(S = 1 \cap X)}{P(X)} = \frac{P(X \cap S = 1)}{P(X)} = \frac{P(X \mid S = 1)P(S = 1)}{P(X)} = \frac{P(X \mid S = 1)P(S = 1)}{P(X \cap S = 1) + P(X \cap S = 0)} = \frac{\frac{1}{2}L}{\frac{1}{2}L + \frac{1}{2}L} = \frac{\frac{1}{2}L}{1L} = \frac{1}{2}$$

Thus, under the assumption that semantic shift does not alter the observable input distribution. $P(X \mid S = 1) = P(X \mid S = 0)$. It follows that $P(S = 1 \mid X = x) = \frac{1}{2}$ for all $x$ in the support of $X$. Therefore, the posterior probability of semantic shift is constant and does not depend on the observed input. This means that $X$ contains no information that can distinguish semantic-shifted samples from clean samples. As a result, any measurable decision rule $f(X)$ must perform no better than a constant classifier, since no input-dependent rule can improve discrimination between $S = 1$ and $S = 0$. Under equal class priors, the Bayes-optimal classifier therefore achieves accuracy $\max\{P(S = 1), P(S = 0)\} = \frac{1}{2}$, which is exactly the chance level. Hence, no classifier based solely on $X$ can distinguish semantic-shifted samples from clean samples better than chance.

**(VI.2) Proposition 2: Benefit of Failure Attribution under Squared Loss**

Let $U = u(X)$ be an uncertainty score and $A = a(X)$ be a failure-attribution representation. Consider predictors of the form $g(U)$ and $h(U, A)$. Under squared loss, the optimal risks satisfy $R^*_{U,A} \leq R^*_U$, with equality if and only if $E[R \mid U, A] = E[R \mid U]$. Moreover, the inequality is strict whenever $A$ contains nonredundant information about $R$ beyond $U$.

**Proof:** Let $X \in X'$ denote the input. Let $R \in \{0, 1\}$ denote the reliability label, where $R = 1$ means the prediction is unreliable. Let $U = u(X)$ denote a standard uncertainty score, and let $A = a(X)$ denote a failure-attribution representation. Consider two function classes, including uncertainty-only predictors of the form $g(U)$, and the attribution-augmented predictors of the form $h(U, A)$.

Under squared loss, we define the optimal prediction risks as $R^*_U = inf_g E[(R - g(U))^2]$ and $R^*_{U,A} = inf_h E[(R - h(U, A))^2]$. $R$ refers to the real value. $g(U)$ and $h(U, A)$ refers to the prediction function. $(R - g(U))^2$ and $(R - h(U, A))^2$ refer to the squared loss, meaning the difference between the predicted value and the real value. $E[]$ refers to the expectation. $inf_g$ and $inf_h$ refers to the functions that minimize the average error.

By the $L^2$ Projection Theorem, which identifies the conditional expectation as the minimizer of the mean squared error ($MSE$), we consider the minimization of the expected squared loss. Since the conditional expectation is the Bayes-optimal predictor, to find the best predictors under the squared loss, we consider the minimization of the expected mean squared error. Since the conditional expectation is the Bayes-optimal predictor with respect to the information available, we define the optimal functions $g^*$ and $h^*$ by projecting $R$ onto the $\sigma$-fields generated by $U$ and $(U, A)$ , respectively.

Therefore, the optimal predictors are $g^*(U) = E[R \mid U]$ and $h^*(U, A) = E[R \mid U, A]$. Hence, $R^*_U = E[(R - E[R \mid U])^2]$, and $R^*_{U,A} = E[(R - E[R \mid U, A]^2]$. By identity, $E[(R - E(R \mid Z))^2] = E[Var(R \mid Z)]$. Since we know that $Var(X) = E[(X - E(X))^2]$, $R^*_U = E[(R - E[R \mid U])^2]$, $R^*_{U,A} = E[(R - E[R \mid U, A)^2]$, and $E[(R - E(R \mid Z))^2] = E[Var(R \mid Z)]$. Therefore, $R^*_U = E[Var(R \mid U)]$ and $R^*_{U,A} = E[Var(R \mid U, A)]$.

Since $Var(R \mid U) = E[Var(R \mid U, A) \mid U] + Var(E[R \mid U, A] \mid U)$, we have got that $E[Var(R \mid U, A) \mid U] \leq Var(R \mid U)$. Then, we can further deduce that $Var(R \mid U, A) \leq Var(R \mid U)$. By taking the expectation, we can get $R^*_{U,A} \leq R^*_U$.

Based on the Law of Total Variance, the general formula is $Var(X) = E[Var(X \mid A)] + Var(E[X \mid A])$. Now, we substitute the $X$ into $(R \mid U)$. Then, we have got: $Var(R \mid U) = E[Var(R \mid U, A) \mid U] + Var[E(R \mid U, A) \mid U]$. By definition of Variance, $Var(R \mid U) = E(R^2 \mid U) - (E[R \mid U])^2$, and based on the Law of Total Expectation, $E[R \mid U] = E[E[R \mid U, A] \mid U]$, we have got that $E[R^2 \mid U] = E[E[R^2 \mid U, A] \mid U]$, and $[E(R \mid U)]^2 = (E[E(R \mid U, A) \mid U])^2$.

Since we know that $Var(R \mid U) = E[E[R^2 \mid U, A] \mid U] - (E[E[R \mid U, A] \mid U])^2$, and $E(R^2 \mid U, A) = Var(R \mid U, A) + (E[R \mid U, A])^2$, then we have got that $Var(R \mid U) = E[Var(R \mid U, A)] + E[[R \mid U, A]^2 \mid U] - (E[E[R \mid U, A] \mid U]^2) =$ $E[Var(R \mid U, A) \mid U] + E[(E[R \mid U, A])^2 \mid U] - (E[E[R \mid U, A] \mid U])^2 =$ $E[Var(R \mid U, A) \mid U] + Var(E[R \mid U, A] \mid U]$.

By taking expectations over $U$, we have got that $R^*_U = R^*_{U,A} + E[Var(E(R \mid U, A) \mid U]$ holds if and only if $Var(E[R \mid U, A \mid U]) = 0$. It means that $E[R \mid U] = E[R \mid U, A]$. The equity occurs when $A$ contributes no additional information about $R$ once $U$ is already known.

**(VI.3) Mathematical Formula of Evaluation Metrics**

Let the test set contain $N$ samples. For the $i$-th sample, we define the true class label as $y_i$, the predicted class label as $\hat{y}_i$, and the ground-truth reliability label (whether the prediction is incorrect) as $r_i = \mathbf{1}(\hat{y}_i \neq y_i)$, where $r_i = 1$ indicates an incorrect prediction, and $r_i = 0$ indicates a correct prediction. Let $\hat{r}_i \in \{0, 1\}$ denote the predicted reliability label, where $\hat{r}_i = 1$ represents predicted unreliable and $\hat{r}_i = 0$ represents predicted reliable. Let $s_i \in [0,1]$ denote the model-produced unreliability score, where larger values indicate higher predicted unreliability. Finally, let $f_i \in \{1,...,K\}$ denote the true failure type and $\hat{f}_i \in \{1,...K\}$ denote the predicted failure type.

- **Reliability Accuracy:** It measures whether the model correctly classifies predictions as reliable or unreliable.

$$\text{Reliability Accuracy} = \frac{1}{N}\sum_{i=1}^{N} \mathbf{1}(\hat{r}_i = r_i)$$

- **Reliability AUROC:** Uses $s_i$ as the unreliability score and $r_i$ as the ground-truth label.

$\text{Reliability AUROC} = \text{AUROC}\left(\{(s_i, r_i)\}_{i=1}^{N}\right)$. This metric evaluates the ranking ability of the model to distinguish correct from incorrect predictions.

- **Error Awareness Rate (EAR):** Among all incorrect predictions, the proportion is correctly flagged as unreliable.

$$\text{Error Awareness Rate (EAR)} = \frac{\sum_i \mathbb{I}(\hat{y}_i \neq y_i \ \wedge \ \hat{r}_i = 1)}{\sum_i \mathbb{I}(\hat{y}_i \neq y_i)}$$

- **Correct Confidence Rate (CCR):** Among all correct predictions, the proportion correctly flagged as reliable.

$$\text{Correct Reliability Rate(CRR)} = \frac{\sum_i \mathbb{I}(\hat{y}_i = y_i \ \wedge \ \hat{r}_i = 0)}{\sum_i \mathbb{I}(\hat{y}_i = y_i)}$$

- **Failure Attribution Accuracy:** Measure how accurately the model predicts the true failure type.

$$\text{Failure Attribution Accuracy} = \frac{1}{N}\sum_{i=1}^{N} \mathbf{1}\left(\hat{f}_i = f_i\right)$$

- **Failure Reason Identification Accuracy on Error Cases:** Among actual prediction errors only, the proportion for which the failure reason is correctly identified.

$$\text{Failure Reason Identification Accuracy on Error Cases} = \frac{\sum_{i=1}^{N} \mathbf{1}\left(\hat{y}_i \neq y_i \wedge \hat{f}_i = f_i\right)}{\sum_{i=1}^{N} \mathbf{1}\left(\hat{y}_i \neq y_i\right)}$$

**(VI.4) Table of Pre-Shift Results**

**Table A**

*Pre-Shift Results*

| Model | Shift | Classification Accuracy | Reliability Accuracy | Error Awareness Rate | Correct Confidence Rate |
|---|---|---|---|---|---|
| 0 | Clean | 0.6754 | 0.8878 | 0.1501 | 0.9060 |
| 0 | Covariate Shift | 0.6648 | 0.9013 | 0.9000 | 0.0980 |
| 0 | Noise Corruption | 0.6428 | 1.0000 | 1.0000 | 0.0000 |
| 0 | Semantic Shift | 0.5302 | 0.9722 | 0.9511 | 0.0091 |
| 1 | Clean | 0.6762 | 0.8033 | 0.3850 | 0.8936 |
| 1 | Covariate Shift | 0.6643 | 0.1917 | 0.3516 | 0.8891 |
| 1 | Noise Corruption | 0.6373 | 0.2182 | 0.3901 | 0.8796 |
| 1 | Semantic Shift | 0.5296 | 0.4640 | 0.5970 | 0.6542 |
| 2 | Clean | 0.7009 | 0.7979 | 0.4027 | 0.8834 |
| 2 | Covariate Shift | 0.7054 | 0.2063 | 0.4103 | 0.8788 |
| 2 | Noise Corruption | 0.6697 | 0.2157 | 0.3937 | 0.8720 |
| 2 | Semantic Shift | 0.4677 | 0.4383 | 0.5527 | 0.6918 |
| 3 | Clean | 0.6754 | 0.7081 | 0.0908 | 0.6115 |
| 3 | Covariate Shift | 0.6648 | 0.3130 | 0.1268 | 0.5930 |
| 3 | Noise Corruption | 0.6428 | 0.2715 | 0.0907 | 0.6279 |
| 3 | Semantic Shift | 0.5302 | 0.0662 | 0.0341 | 0.9053 |
| 4 | Clean | 0.6379 | 0.9115 | 0.1166 | 0.9275 |
| 4 | Covariate Shift | 0.6464 | 0.9160 | 0.9159 | 0.0840 |
| 4 | Noise Corruption | 0.5899 | 1.0000 | 1.0000 | 0.0000 |
| 4 | Semantic Shift | 0.4682 | 0.9851 | 0.9809 | 0.0103 |

*Note.* 0 refers to the Uncertainty-Only Model. 1 refers to the MC Dropout Model. 2 refers to the Deep Ensemble Model. 3 refers to the Energy-Based Score Model. 4 refers to the Self-Diagnosing + Failure Attribution Model.

**VII. Reference**

Gal, Y., & Ghahramani, Z. (2015). Dropout as a Bayesian Approximation: Representing Model Uncertainty in Deep Learning. *ArXiv (Cornell University)*. https://doi.org/10.48550/arxiv.1506.02142

Hendrycks, D., & Gimpel, K. (2018). A Baseline for Detecting Misclassified and Out-of-Distribution Examples in Neural Networks. *ArXiv (Cornell University)*. https://doi.org/10.48550/arXiv.1610.02136

Hendrycks, D., & Dietterich, T. G. (2019). Benchmarking Neural Network Robustness to Common Corruptions and Perturbations. *ArXiv (Cornell University)*. https://doi.org/10.48550/arxiv.1903.12261

Hendrycks, D., Mazeika, M., & Dietterich, T. (2019). Deep Anomaly Detection with Outlier Exposure. *ArXiv (Cornell University)*. https://arxiv.org/abs/1812.04606

Krizhevsky, A. (2009). *CIFAR-10 and CIFAR-100 datasets*. Toronto.edu. https://www.cs.toronto.edu/~kriz/cifar.html

Lakshminarayanan, B., et al. (2016). Simple and Scalable Predictive Uncertainty Estimation using Deep Ensembles. *ArXiv (Cornell University)*. https://doi.org/10.48550/arxiv.1612.01474

Lee, K., Lee, K., Lee, H., & Shin, J. (2018). A Simple Unified Framework for Detecting Out-of-Distribution Samples and Adversarial Attacks. *ArXiv (Cornell University)*. https://doi.org/10.48550/arXiv.1807.03888

Liang, S., Li, Y., & Srikant, R. (2020). Enhancing The Reliability of Out-of-distribution Image Detection in Neural Networks. *ArXiv (Cornell University)*. https://doi.org/10.48550/arXiv.1706.02690

Liu, W., Wang, X., Owens, J. D., & Li, Y. (2020). Energy-based Out-of-distribution Detection. *ArXiv (Cornell University)*. https://doi.org/10.48550/arXiv.2010.03759

Ovadia, Y., Fertig, E., Ren, J., Nado, Z., Sculley, D., Nowozin, S., Dillon, J. V., Lakshminarayanan, B., & Snoek, J. (2019). Can You Trust Your Model's Uncertainty? Evaluating Predictive Uncertainty Under Dataset Shift. *ArXiv (Cornell University)*. https://doi.org/10.48550/arXiv.1906.02530

Quiñonero-Candela, J. (2009). *Dataset Shift In Machine Learning*. Mit Press.

Wilson, A. G., & Izmailov, P. (2020). Bayesian Deep Learning and a Probabilistic Perspective of Generalization. *ArXiv (Cornell University)*. https://doi.org/10.48550/arXiv.2002.08791

Yang, J., Zhou, K., Li, Y., & Liu, Z. (2021). Generalized Out-of-Distribution Detection: A Survey. *ArXiv (Cornell University)*. https://doi.org/10.48550/arXiv.2110.11334